\documentclass[11pt]{article}

\usepackage[margin=1in]{geometry}
\usepackage{times}
\usepackage{microtype}
\usepackage{hyperref}
\hypersetup{colorlinks=true, linkcolor=blue!60!black, citecolor=blue!60!black, urlcolor=blue!60!black}
\usepackage{array}
\usepackage[table]{xcolor}
\usepackage{soul}
\usepackage{amsmath}
\usepackage{graphicx}
\usepackage[backend=bibtex, style=numeric, maxbibnames=4, minbibnames=4, sorting=nyt]{biblatex}
\usepackage{titlesec}
\titlespacing*{\paragraph}
  {0pt}    
  {0.5ex}  
  {0.5em}  

\title{\bfseries Agency as a New Organizing Abstraction for Computer Science}

\author{
Ian Foster\\
\small Argonne National Laboratory and The University of Chicago\\
\small \texttt{foster@anl.gov}
}

\date{}

\begin{document}
\maketitle
\let\thefootnote\relax\footnotetext{Expanded version of a paper appearing in \textit{Communications of the ACM}, 2026.}

\vspace{-1em}


\vspace{1em}

\begin{abstract}
Advances in large-scale machine learning have enabled agentic AI systems: persistent, autonomous computational entities that pursue goals, initiate actions, and interact with open environments. These systems are often framed as applications of foundation models. We argue that this framing is incomplete: agency is also a candidate new \textit{organizing abstraction} for computer science. What is new is not the concept of an agent, nor runtime policy generation per se, but its general-purpose, open-ended form, in which systems can synthesize courses of action from goals, observations, and memory, over tools and environments for which no complete task-specific model was specified in advance. Thus agency becomes a systems-level abstraction as well as an AI modeling concept. Where the process abstraction organizes computation around executing programs, agency organizes it around pursuing goals under bounded resources: \textit{a process executes a program; an agent pursues a goal}. From this perspective, the recurring failures of deployed agentic systems (runaway resource consumption, behavior redirected by adversarial content, opaque continuous operation) are symptoms of structural abstraction gaps, not isolated engineering defects. We examine these gaps across six dimensions (execution, interfaces, correctness, oversight, coordination, and persistence) and argue that closing them requires coordinated advances in systems, programming languages, theory, and human--computer interaction.
\end{abstract}

\section{From Models to Agents}

The surge of interest in artificial intelligence (AI) has been driven largely by advances in machine learning models, particularly large neural networks trained on massive datasets. These models have achieved striking capabilities in perception, language, and pattern recognition, reshaping both research and practice. 

At the same time, an increasing number of deployed AI systems are no longer best understood as isolated inference engines. Instead, they take the form of \emph{agents\/}: computational entities that persist over time, initiate actions, pursue goals, and interact continuously with their environments. Classical AI research has long defined agents as entities that perceive and act within an environment \cite{russell_norvig}, and multiagent systems research has explored autonomy and interaction for decades \cite{wooldridge_agents}. What is new is not the conceptual notion of an agent, nor runtime policy generation per se, but its general-purpose, open-ended form: foundation models now enable systems to synthesize courses of action from goals, observations, and memory, over tools and environments for which no complete task-specific model was specified in advance.

\begin{figure}[t]
\centering
\includegraphics[width=0.95\textwidth]{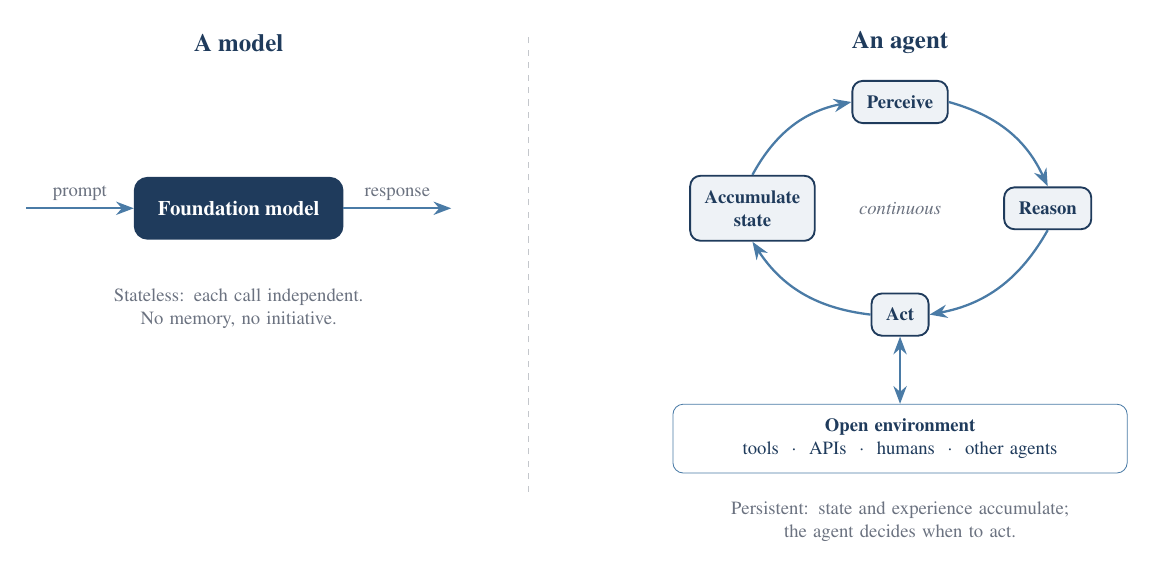}
\caption{From models to agents. A foundation model is a stateless inference engine that maps prompts to responses; an agent embeds such a model in a persistent perceive--reason--act loop, accumulating state and interacting continuously with an open environment.}
\label{fig:model_vs_agent}
\end{figure}

Three developments illustrate this new capability.
Language models can learn to select and invoke external APIs~\cite{schick_toolformer}: a tool-use capability that, embedded in a larger architecture, transforms stateless inference engines into systems that take consequential actions.
Multi-agent frameworks enable AI agents to collaborate on complex tasks through extended conversations, persisting state across interactions that may span hours or days~\cite{wu_autogen}.
Agents that interleave reasoning and action can execute multi-step workflows, invoking computational tools and interpreting results without step-by-step human guidance \cite{yao2022react}.
Figure~\ref{fig:model_vs_agent} contrasts the two modes of operation: a stateless foundation model that maps prompts to responses, and an agent that embeds such a model in a persistent loop of perception, reasoning, and action.

More than a new application layer, the transition from stateless model to persistent agent is a reorganization of computation that challenges foundational assumptions across operating systems, programming languages, verification, and human--computer interaction. These challenges motivate treating agency not merely as an application-level concern but as a candidate organizing abstraction: a framing that, as we explain in more detail below, reveals the difficulties agentic systems encounter (runaway resource consumption, ambiguous lifecycles, opaque failures) as structural rather than isolated engineering problems.

The thesis of this paper is as follows:

\begin{quote}
\itshape A process executes a program; an agent pursues a goal. Whereas a process's behavior is principally determined by code specified before execution, an agent generates and revises courses of action at runtime from goals, observations, memory, and available capabilities. Agency therefore shifts the organizing problem from controlling execution to controlling goal-directed behavior under bounded resources and authority.
\end{quote}

More generally, while computer science has powerful abstractions for controlling the execution of computations whose behavior has been specified in advance, it lacks comparably mature abstractions for controlling computations that decide, at runtime, what behavior to generate in pursuit of goals. That gap explains why the concerns developed in this paper (budgets, contracts, behavioral correctness, oversight, provenance, memory, delegation) cluster together: each arises because future behavior is generated rather than prescribed. Table~\ref{tab:transitions} previews the six dimensions along which we develop this argument.

\begin{table}[t]
\centering
\caption{Six dimensions along which agency extends existing abstractions. In each, prior work addresses the right-hand concern in specialized settings; agency makes it a general and intrinsic requirement (Section~\ref{sec:reveals}).}
\vspace{1ex}
\label{tab:transitions}
\begin{tabular}{lll}
\hline
\textbf{Dimension} & \textbf{Conventional organizing focus} & \textbf{Agency makes first-class} \\
\hline
Execution & Resource limits & Resource budgets \\
Interfaces & Protocols & Contracts \\
Correctness & Functional correctness & Behavioral correctness \\
Oversight & Commands & Attention allocation \\
Coordination & Scheduling & Negotiation \\
Persistence & State & Memory \\
\hline
\end{tabular}
\end{table}

The remainder of this paper is organized as follows. Section~\ref{sec:abstractions} examines the role that organizing abstractions have played in computer science, tracing a lineage from procedures to foundation models and identifying the gap that agency fills. Section~\ref{sec:agentic} characterizes agentic systems in terms of four defining properties, the key distinction (open-ended policy generation), theoretical grounding in bounded rationality, and a working definition. Section~\ref{sec:failures} documents recurring failure modes of deployed agentic systems and the abstraction gaps they expose. Section~\ref{sec:reveals} develops the problems that the agency abstraction reveals, across six dimensions: execution, interfaces, correctness, oversight, coordination, and persistence. Section~\ref{sec:implications} draws implications for computer science research, Section~\ref{sec:related} positions this work relative to prior traditions, and Section~\ref{sec:conclusion} concludes.

\emph{A note on scope: agentic AI raises profound questions of governance, accountability, and social impact~\cite{singapore_consensus}. These questions are important, and better abstractions alone will not resolve them; but they demand treatment of their own, and this article confines itself to technical challenges amenable to computer science research.}

\section{Organizing Abstractions in Computer Science}
\label{sec:abstractions}

Before discussing whether agency merits recognition as an organizing abstraction, we first ask: what \textit{is} an organizing abstraction, and what role have such abstractions played in the development of computer science?

\subsection{What Is an Organizing Abstraction?}

Computer science has progressed through a series of conceptual innovations that do more than solve individual problems: they restructure entire subfields by providing new ways to think about, build, and reason about computational systems. We call these \textit{organizing abstractions}: conceptual frameworks that make certain problems easy to state, tractable to solve, and natural to compose.

An organizing abstraction is not just a useful technique or a clever implementation strategy. It is a way of carving up the computational world that reveals structure previously hidden. Thus, for example, the \emph{procedure} made control flow explicit and composable; the \emph{object} bundled state with behavior in ways that enabled new forms of modularity; and the \emph{process} gave operating systems a unit of isolation, scheduling, and resource accounting.

Each organizing abstraction also has blind spots: problems that it renders invisible or intractable. Procedures obscure state; early object models assume co-location; the process abstraction has no semantics for goals or for the value of continued execution. Recognizing these limitations is essential to understanding when a new abstraction may be needed.

\subsection{A Lineage of Organizing Abstractions}

\begin{figure}[t]
\centering
\includegraphics[width=0.62\textwidth]{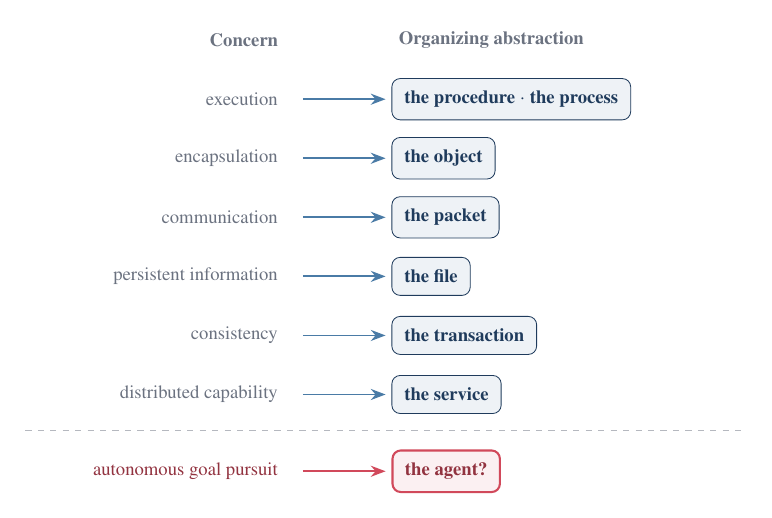}
\caption{Organizing abstractions address distinct concerns and are complements rather than successors. We ask whether autonomous goal pursuit is a concern of comparable generality, and the agent its organizing abstraction.}
\label{fig:lineage}
\end{figure}

The history of computer science can be read, in part, as the accumulation of organizing abstractions, each addressing a distinct concern. The procedure and the process organize execution, the object organizes encapsulation, the packet organizes communication, the file organizes persistent information, the transaction organizes consistency, and the service organizes distributed capability (Figure~\ref{fig:lineage}). These abstractions are complements, not successors: the transaction did not replace the file, nor the packet the procedure. The question this paper asks is whether \textit{autonomous goal pursuit} is a concern of comparable generality that is currently underrepresented in this canon. The agent, on this reading, would not replace the process any more than the transaction replaced the file; it would organize a dimension of computation that existing abstractions leave implicit. In what follows we examine the abstractions most relevant to that question.

\paragraph{Procedures.}
The procedure organized computation around named sequences of instructions with local scope and explicit control flow. Structured programming, in the 1960s, replaced ad-hoc control flow with disciplined composition of procedures~\cite{dijkstra_structured}, making programs easier to write, read, and verify. The procedure abstraction made modularity and reuse tractable, but provided no mechanism for encapsulating state with behavior: a limitation that motivated subsequent abstractions.

\paragraph{Objects.}
Object-oriented programming, emerging from Simula~\cite{dahl_simula} and refined in Smalltalk~\cite{kay_smalltalk}, bundled state and behavior into encapsulated units communicating via message passing. Objects enabled new forms of modularity through inheritance and polymorphism, and provided a natural vocabulary for modeling domains where entities have identity and behavior. However, early object models largely assumed synchronous, co-located execution: assumptions that distributed systems would later stress.

\paragraph{Processes.}
The process abstraction, central to operating systems since the 1960s, organized computation around isolated execution contexts with their own address spaces, managed by a kernel that handles scheduling, resource allocation, and protection~\cite{dijkstra_the}. Processes made concurrency and protection tractable: multiple computations could proceed independently without interfering, and failures could be contained. Processes can fork children, run as daemons, and terminate themselves; the important point is not that the mechanics forbid self-direction, but that the abstraction itself has no semantics for goals, for authority tied to goals, or for the value of continued execution. Resource bounds are imposed from outside, and termination is simply an event, never a decision to be weighed. These gaps, largely invisible for decades, become consequential for agentic systems.

\paragraph{Actors.}
The actor model, introduced by Hewitt, Bishop, and Steiger~\cite{hewitt_actors}, organized concurrent computation around isolated entities communicating exclusively via asynchronous message passing. Actors eliminated shared mutable state, simplifying reasoning about concurrent systems. Each actor processes messages sequentially, may create new actors, and may send messages to known addresses. 
The actor model influenced subsequent distributed systems and programming languages, from Erlang~\cite{armstrong_erlang} to the Akka framework for the JVM.
Critically for our purposes, the actor abstraction organizes behavior around message handling rather than goal pursuit: its semantics specify how actors respond to messages, not how they formulate and pursue goals.

\paragraph{Services.}
Service-oriented architecture organized distributed computation around loosely coupled components exposing well-defined interfaces over network protocols~\cite{erl_soa}. Services enabled interoperability across organizational and technological boundaries, and microservices refined this model for fine-grained decomposition~\cite{newman_microservices}. The abstraction's reach extended beyond enterprise computing: \textit{service-oriented science} proposed exposing data, instruments, and analyses as composable network services, reorganizing the practice of research itself~\cite{foster_sos}. The canonical service abstraction assumes request-response semantics: a client initiates, a service responds, and control returns to the client. Statelessness (or session-bounded state) simplifies scaling and failure recovery. These assumptions embed a model of computation in which initiative and orchestration reside outside the service itself: in the case of science, with the human researcher who composes and invokes them.

\paragraph{Dataflows.}
Dataflow models organized computation around directed graphs of transformations over data~\cite{johnston_dataflow}. From early dataflow architectures through modern systems like MapReduce, Spark, TensorFlow, and streaming platforms, the dataflow abstraction made data parallelism and pipeline composition tractable. Dataflows excel when the structure of computation is known at design time; they are less well suited to computations whose structure emerges dynamically.

\paragraph{Virtual machines and containers.}
Virtualization abstractions, from the virtual machines of 1960s IBM mainframes~\cite{creasy_vm370} through their modern revival~\cite{barham_xen} to containers, organized deployment around portable, isolated execution environments with controlled resource access. These abstractions revolutionized infrastructure management, enabled cloud computing, and were proposed early on as a foundation for grid computing across administrative domains~\cite{figueiredo_vm_grid}. However, they inherit the process model's silences: nothing in the abstraction connects a workload to a goal, ties its resource consumption to the value of that goal, or weighs the cost of terminating it.

\paragraph{Grids and virtual organizations.}
Grid computing organized large-scale distributed computation around coordinated resource sharing across institutional boundaries~\cite{foster_anatomy}. The \textit{virtual organization} abstraction addressed a problem that resonates with agentic systems: collaboration among autonomous entities with independent administrative control, heterogeneous trust relationships, and dynamically negotiated policies. Its mechanisms for establishing trust, delegating credentials, and negotiating resource access reappear in multi-agent systems.

Early work recognized the need to integrate Grid infrastructure with agent-based reasoning~\cite{foster_brain_brawn}, characterizing Grids as ``all brawn and no brain'' (robust mechanisms for secure resource sharing, limited capacity for autonomous decision-making) and agent systems as the reverse, and proposing convergence between the two. The research agenda articulated in that work, spanning trust negotiation, negotiation-based contracts, emergent behavior, and human--computer collaboration, anticipates many of the challenges identified in this paper. For two decades, however, the convergence remained largely aspirational: Grid participants executed \textit{predefined} policies, and agent systems reasoned flexibly only within narrow domains. We argue that foundation models supply the missing ingredient---a general-purpose policy generator that can reason about goals, plans, and actions across diverse domains---and that its arrival is what makes the current moment qualitatively different from earlier visions of agent--Grid integration.

\paragraph{Foundation models.}
Most recently, foundation models---large neural networks trained on broad data and adaptable to diverse downstream tasks~\cite{bommasani_foundation}---have emerged as a powerful abstraction for general-purpose inference: a single model can be prompted or fine-tuned to perform classification, generation, translation, reasoning, and many other tasks. Foundation models provide flexible capabilities but, in their basic form, are stateless inference engines: they map inputs to outputs without persistence, goals, or autonomous action. The transition from foundation models to agentic systems is the transition from capabilities to agency.

\subsection{The Gap That Agency Fills}

Each abstraction in this lineage solved important problems while leaving others unaddressed. We can now identify a specific gap that motivates treating agency as a candidate organizing abstraction: none of the abstractions surveyed above natively supports computational entities that \textit{persist} with accumulated state, \textit{initiate} actions autonomously, \textit{pursue goals} that may be underspecified or conflicting, and \textit{operate in open environments} that cannot be fully specified in advance. The actor model comes closest, but its semantics specify how an actor responds to messages, not how it formulates and pursues goals; the other abstractions locate goals, initiative, and control outside the computational entity altogether.

The question, then, is whether the combination of these four properties is sufficiently coherent and consequential to merit a new organizing abstraction, or whether it can be adequately handled by composing or extending existing ones. The remainder of this paper argues for the former: that agency reveals a set of related problems that become visible and tractable only when we take agency itself as the organizing concept.

\section{What Makes a System Agentic?}
\label{sec:agentic}

We use the term \textit{agency} to denote persistent, goal-directed computation in which courses of action are generated dynamically, under bounded resources, in an incompletely specified environment that includes humans. This section develops that characterization in four steps: the operational properties that agentic systems exhibit (Section~\ref{sec:properties}), the mechanism that distinguishes contemporary systems from their predecessors (Section~\ref{sec:policygen}), theoretical grounding (Section~\ref{sec:bounded}), and a working definition (Section~\ref{sec:definition}).

\subsection{Four Properties of Agentic Systems}
\label{sec:properties}

Agentic systems are \textbf{persistent}, operating over extended periods and accumulating state, context, and experience. Unlike stateless services or batch jobs, they maintain continuity that influences future behavior. This persistence complicates resource management, failure handling, and reasoning about long-term effects, particularly when systems must be upgraded or audited while in operation.

Agentic systems are \textbf{autonomous}. They decide when to act rather than responding solely to external requests. This autonomy, though partial and constrained, fundamentally shifts computation from reactive to proactive.

Agentic systems are \textbf{goal-directed}. Their behavior is shaped by objectives that may require planning, monitoring, adaptation, and revision over time. Goals are often under-specified or conflicting, forcing agents to make tradeoffs rather than execute fixed procedures.

Finally, agentic systems operate in \textbf{open environments}, interacting with APIs, tools, physical instruments, humans, and other agents. What makes an environment \textit{open} in our sense is not mere unpredictability, which any distributed system faces, but that the agent must interpret semantically described observations and capabilities whose relevance and effects are not captured by a complete task-specific model. Agents' actions are frequently consequential, consuming real resources and producing effects that may be difficult or impossible to undo.

\begin{figure}[t]
\centering
\includegraphics[width=0.9\textwidth]{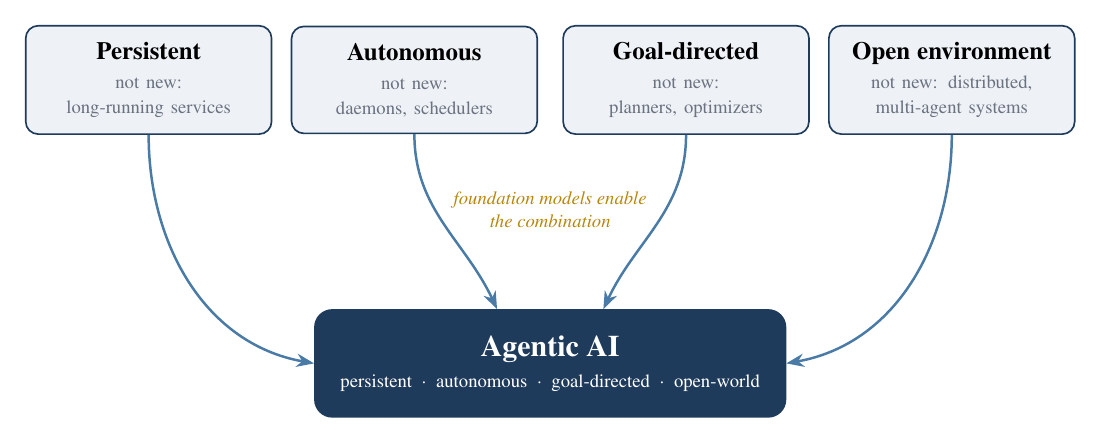}
\caption{Four properties characterize agentic operation. Each has precedent in earlier systems; what is new is their combination in a single general-purpose system, enabled by foundation models. The deeper distinction lies in how behavior is produced (Section~\ref{sec:policygen}).}
\label{fig:four_properties}
\end{figure}

Individually, none of these properties is new: long-running services persist, daemons act without prompting, planners pursue objectives, and distributed systems operate in open settings. What distinguishes contemporary agentic systems is the combination of all four properties in a single general-purpose system, enabled by foundation models that support flexible behavior across diverse domains (Figure~\ref{fig:four_properties}). Yet the combination alone does not fully capture the difference; the deeper distinction lies in how these systems produce behavior (Section~\ref{sec:policygen}).

\subsection{The Key Distinction: Policy Generation}
\label{sec:policygen}

A skeptical reader might observe that many existing systems exhibit some or all of these properties. Kubernetes operators persist and act autonomously to reconcile cluster state~\cite{burns_borg}. Automated trading systems pursue profit objectives in open markets. Autonomous spacecraft operate for years, making decisions in environments where communication delays preclude human control~\cite{muscettola_remote_agent}. Adaptive control systems adjust behavior in response to changing conditions.

Why, then, should we consider agency a new abstraction rather than an incremental extension of existing ones?

The answer lies in a distinction that has remained implicit in much discussion of agentic AI: the distinction between \textit{policy execution} and \textit{policy generation}.

A Kubernetes operator executes a predefined reconciliation policy: if the observed state diverges from the desired state, apply a known sequence of corrective actions. The policy, the mapping from observations to actions, is specified by a human developer and fixed at deployment time. The operator may be autonomous in \textit{execution}, but its behavior is determined by a policy that exists prior to any particular situation it encounters.

Similarly, an automated trading system executes strategies defined by quantitative analysts. An autonomous spacecraft follows mission plans with predefined contingency branches. An adaptive controller adjusts parameters within a control law whose structure is fixed.

Learning systems complicate this distinction. A reinforcement-learning agent can learn policies that produce behaviors not explicitly anticipated by its designers, and a foundation model is itself a fixed parameterized function at inference time. The distinction we emphasize is therefore not simply novelty of behavior. It is that contemporary foundation-model agents can construct and revise courses of action over semantically described tools and environments without requiring designers to supply an explicit task-specific state-transition model or to enumerate the relevant behaviors in advance. Given a goal expressed in natural language and access to an environment, such an agent \textit{generates} a course of action that did not exist before the goal was specified, reasoning about the goal, the environment, and the consequences of potential actions rather than selecting from a predefined repertoire.

This is \textbf{open-ended policy generation}, the capability highlighted in the introduction: the operational policy is not fully specified at design time; instead, the system synthesizes context-dependent courses of action at runtime from goals, observations, memory, and available capabilities. Any implemented agent remains bounded by its action vocabulary, its interfaces, its permissions, and its compute; but within those bounds it may invoke tools in combinations never foreseen, pursue subgoals never specified, and adapt its approach in ways that emerge from its reasoning rather than from explicit programming.

The distinction is best understood as a continuum rather than a binary: \textit{fixed procedure} $\rightarrow$ \textit{parameterized policy} $\rightarrow$ \textit{planner or synthesizer} $\rightarrow$ \textit{open-ended policy generator}. Each step moves more of the determination of behavior from design time to runtime. Agency names the regime at the far end of this continuum, where behavior is determined principally at runtime.

This shift changes what pre-deployment analysis can accomplish. When a policy is fixed at design time, it can be inspected, tested, and analyzed before deployment: however complex the resulting behavior, the mapping from situation to response exists in advance. When the policy is synthesized at runtime,     the particular policy does not exist in advance; the subject of verification shifts from a policy to a policy \textit{generator}.

\paragraph{Why is this not simply planning?} Readers familiar with automated planning may object that synthesizing behavior at runtime is a mature technology: classical planners, hierarchical task networks, program synthesis, and reinforcement learning all construct behavior rather than selecting it from a fixed repertoire~\cite{ghallab_planning,gulwani_synthesis}. Indeed, the Remote Agent synthesized plans in response to goals and changing spacecraft conditions a quarter century ago~\cite{muscettola_remote_agent}. The difference lies at the \textit{action--model boundary}. Classical planning presupposes an explicit model: a formal state representation, an action vocabulary with transition semantics, and a goal representation, all constructed by engineers before the planner runs. Contemporary agents construct courses of action over tools and semantic environments whose relevant state and action models are incomplete, discovered dynamically, or represented only implicitly in a foundation model. The planner's hard problem was search within a specified model; the agent's hard problem includes deciding, while acting, what the model \textit{is}: which observations matter, what a tool will do, whether an action is reversible. This is why the challenges developed in this paper center on constraining, auditing, and supervising generated behavior rather than on search efficiency.

\subsection{Theoretical Grounding: Bounded Rationality}
\label{sec:bounded}

The shift from policy execution to policy generation invites a theoretical framing. We suggest that Herbert Simon's concept of \textit{bounded rationality}~\cite{simon_bounded} provides useful foundations.

Simon argued that rational decision-making must be understood relative to the constraints under which decisions are made. Real agents, whether human or artificial, operate with limited information, limited computational resources, and limited time. They cannot optimize over all possible futures; instead, they \textit{satisfice}: they seek solutions that are good enough given their constraints, rather than solutions that are globally optimal.

This framing fits agentic systems well. An agent pursuing a goal generally cannot compute an optimal policy over all possible futures; it must instead generate a course of action that appears adequate given its current understanding and available resources.

Bounded rationality also suggests how we might think about correctness for agentic systems. Classical functional correctness typically separates whether a result satisfies a specification from the resources required to obtain it; real-time systems and anytime algorithms are notable exceptions, but the separation is the default. For agents, by contrast, the quality of a decision may itself need to be judged relative to the information, time, and resources available when the decision was made: an agent behaves correctly if it makes reasonable decisions given what it knows and what it can do.

This perspective connects to formal notions, among them regret bounds, safety envelopes, and satisficing criteria, to which we return in our discussion of verification (Section~\ref{sec:correctness}).

Bounded rationality also does more than motivate correctness criteria. An agent continuously decides not only \textit{what action to take}, but \textit{how much computation, information, money, risk, and human attention to spend deciding what action to take}. Managing its own bounds is part of its computation. This observation links much of what follows: resource budgets, escalation to human supervisors, tool selection, delegation, termination decisions, and graceful degradation are all instances of an agent allocating bounded resources to the pursuit of a goal.

We noted that the process abstraction organizes computation around \textit{executing programs}: the transformation of inputs to outputs according to code specified in advance. We propose that the agency abstraction organizes computation around \textit{pursuing goals under bounded resources}. While a process executes a program, an agent pursues a goal, by generating and revising courses of action under constraints of information, resources, authority, and time.

\subsection{A Working Definition}
\label{sec:definition}

The preceding discussion suggests a working definition suitable for computer science research:

\begin{quote}
An \textbf{agent} is a computational entity characterized by a tuple $(G, \Pi, M, E)$ where:
\begin{itemize}
    \item $G$ is a \textit{goal specification} (possibly partial, possibly evolving) that defines what the agent is trying to achieve;
    \item $\Pi$ is a \textit{policy generator} that dynamically determines courses of action from goals, observations, memory, and available capabilities; its output may be a plan, a single action, a contingent policy, or a revised subgoal;
    \item $M$ is \textit{memory\/}: accumulated state that persists across interactions and influences future behavior;
    \item $E$ is an \textit{environment interface\/}: the set of actions available to the agent and the observations it can receive.
\end{itemize}
\end{quote}

We offer $(G, \Pi, M, E)$ as a \textit{conceptual decomposition}, not a formal computational model: we deliberately do not fix the signature of $\Pi$, the representation of observations and histories, or the semantics of goal satisfaction. A rigorous formal model, one specifying how policies, histories, budgets, and constraints interact, is an important research goal in its own right; the decomposition suffices here to locate the abstraction gaps.

The key distinction is $\Pi$: a general-purpose generator that determines courses of action at runtime without requiring a complete task-specific model to have been specified in advance. A process has a program; an agent has a policy generator. A service has an API implementation; an agent has the capacity to decide what actions to take in pursuit of goals it has been given.

The definition is deliberately restrictive. A stateless language model is not an agent under it: it lacks memory $M$ and does not pursue goals $G$ across interactions. Nor is a Kubernetes operator: its policy is fixed, not generated. A robot following a predefined trajectory is likewise not an agent: its course of action is specified in advance rather than generated at runtime in pursuit of a goal.

Conversely, a system that maintains context across conversations, pursues user-specified objectives through dynamically chosen actions, and adapts its approach based on feedback \textit{is} an agent under this definition, and the challenges identified in this paper apply.

Beyond the tuple itself, two aspects of agency require treatment that conventional systems abstractions do not provide. The first is \textit{authority}. In most deployed settings, an agent does not act in its own right: it acts under authority ultimately derived from a principal. Capability and authority are distinct: that an agent \textit{can} send an email, spend money, or delete a database does not mean it is \textit{authorized} to do so in pursuit of every goal. Processes carry a user identity, and access control mediates individual operations; but no existing abstraction represents the scoped, goal-dependent, delegable authority that agents require. Authority is deliberately not part of the descriptive tuple: it is a property of the relationship between an agent and its principal, not of the agent's internal computational state. Classical work on capability systems, least privilege, and the confused-deputy problem~\cite{dennis_capabilities,saltzer_protection,hardy_deputy} applies directly; we return to authority in our discussions of contracts and delegation (Sections~\ref{sec:contracts} and~\ref{sec:coordination}).

The second is the \textit{goal} itself. Traditional systems manipulate values, messages, requests, jobs, and transactions; an agent manipulates something more unusual: a desired future state, often expressed in natural language and incompletely specified. Goals raise semantic questions that conventional program abstractions do not make first class. Who may create or modify a goal? May an agent create subgoals, and which constraints does a subgoal inherit? When is a goal satisfied, and when should it be abandoned? May an observation alter a goal, or only beliefs about how to achieve it? In short: existing abstractions make \textit{execution} first class; agency makes \textit{goals, and the controlled pursuit of goals,} first class.

\begin{figure}[htb]
\centering
\includegraphics[width=0.42\textwidth]{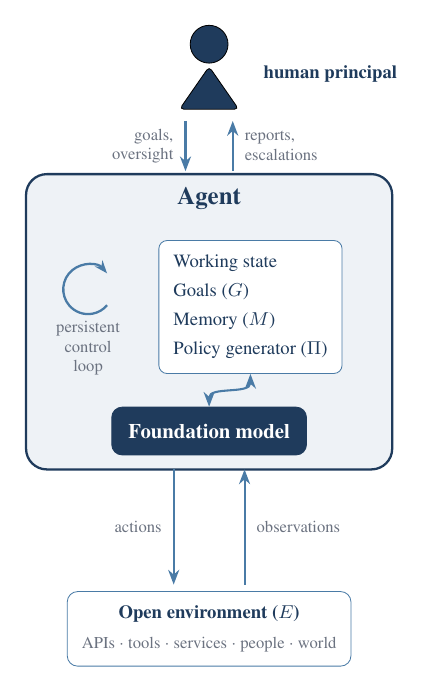}
\caption{Architecture of an agentic system. The agent comprises ephemeral working state, goals ($G$), persistent memory ($M$), and a policy generator ($\Pi$). A foundation model provides general-purpose reasoning for policy generation. The agent interacts with an open environment ($E$) through actions and observations, running a persistent control loop under human oversight.}
\label{fig:agent_architecture}
\end{figure}

Figure~\ref{fig:agent_architecture} illustrates this architecture. The agent maintains ephemeral working state alongside the goals, memory, and policy generator of our working definition. A foundation model provides the reasoning capability that enables open-ended policy generation. The agent runs a persistent control loop, interacting with an environment through actions and observations. A human principal provides goals and oversight, but does not direct each action.

\section{Where Current Abstractions Fall Short}
\label{sec:failures}

Deployments of agentic systems expose a set of recurring failure modes that are not well addressed by existing computational abstractions.

\begin{figure}[t]
\centering
\includegraphics[width=\textwidth,trim=0.4cm 0.4cm 0.4cm 0.4cm,clip]{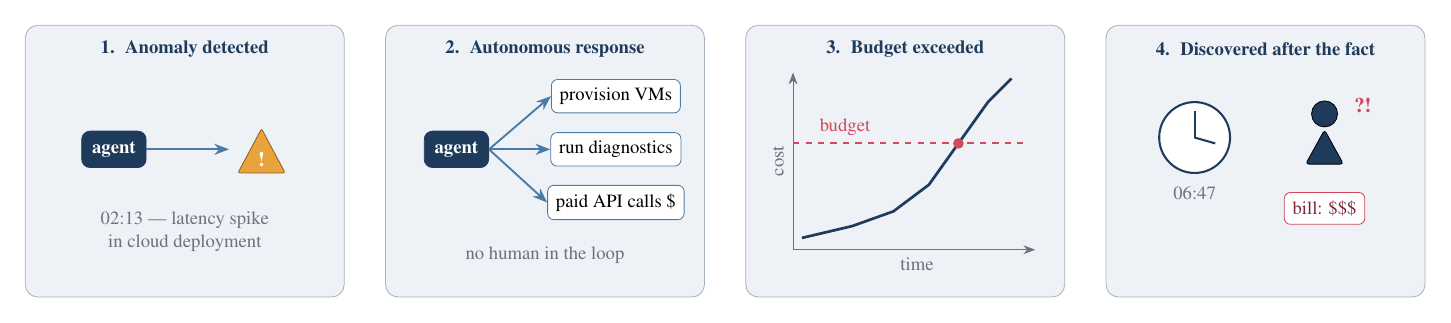}
\caption{A runaway-agent scenario. An agent monitoring a cloud deployment detects an anomaly and responds autonomously; the cost of its actions becomes visible only after budgets have been exceeded. Reactive resource control arrives too late; agentic systems require proactive constraint enforcement.}
\label{fig:runaway}
\end{figure}

A first class of failure concerns runaway resource consumption. Consider an agent tasked with monitoring a cloud deployment. When it detects an anomaly, it may autonomously provision resources, invoke diagnostic tools, or roll back changes (Figure~\ref{fig:runaway}).
Current container and process abstractions assume externally defined resource bounds. An agent that autonomously spawns subprocesses, allocates memory, or invokes paid APIs can exceed budgets in ways that become visible only after significant cost has been incurred~\cite{kapoor_agents_matter,agent_contracts,eet_termination}.

Nor is this failure mode specific to cloud operations. A scientific agent that drives instruments and launches simulation campaigns, a personal assistant that acts across email, calendars, and purchases, a software-engineering agent that edits and deploys code, and a robot that manipulates the physical world all exhibit the same pattern: autonomous initiation of consequential actions, costs and effects that become visible only after the fact, and no existing abstraction positioned to enforce constraints before actions occur. The recurrence of the same gaps across such different domains is evidence that the gaps are structural.

\begin{figure}[t]
\centering
\includegraphics[width=0.95\textwidth]{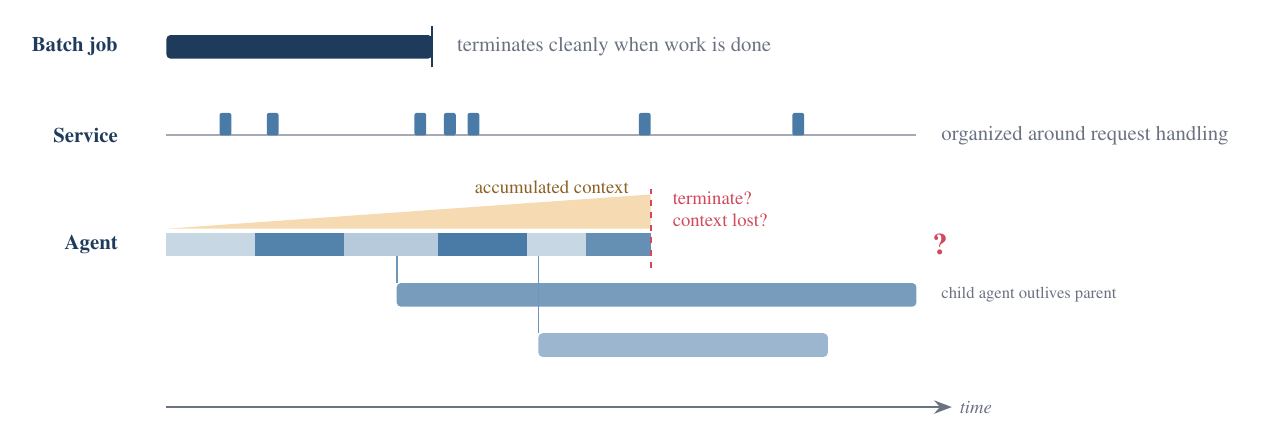}
\caption{Lifecycle mismatch. Batch jobs run toward termination; services are organized around periods of request handling; agents blur these boundaries: intermittently active, spawning children that outlive their parent, and accumulating context that makes termination costly.}
\label{fig:lifecycles}
\end{figure}

A second difficulty concerns lifecycle management. Agentic systems do not fit cleanly into existing notions of process startup, execution, and termination.
Conventional lifecycle abstractions provide recognizable phases: batch jobs run toward termination, while services are organized around periods of request handling. Agentic systems blur these boundaries. An agent may be ``running'' in the sense of maintaining state and awaiting opportunities to act, without actively consuming cycles. It may spawn child agents that outlive their parent, each pursuing independent goals with its own accumulated context, unlike forked processes, which inherit fixed execution state. It may accumulate context that makes termination costly in terms of lost learning.
When should such a system be terminated? How should its accumulated state be preserved or transferred?
Current process models provide no principled answers~\cite{agentops_survey}.
Figure~\ref{fig:lifecycles} contrasts the agent lifecycle with the batch and service lifecycles that existing abstractions assume.

\begin{figure}[t]
\centering
\includegraphics[width=0.95\textwidth]{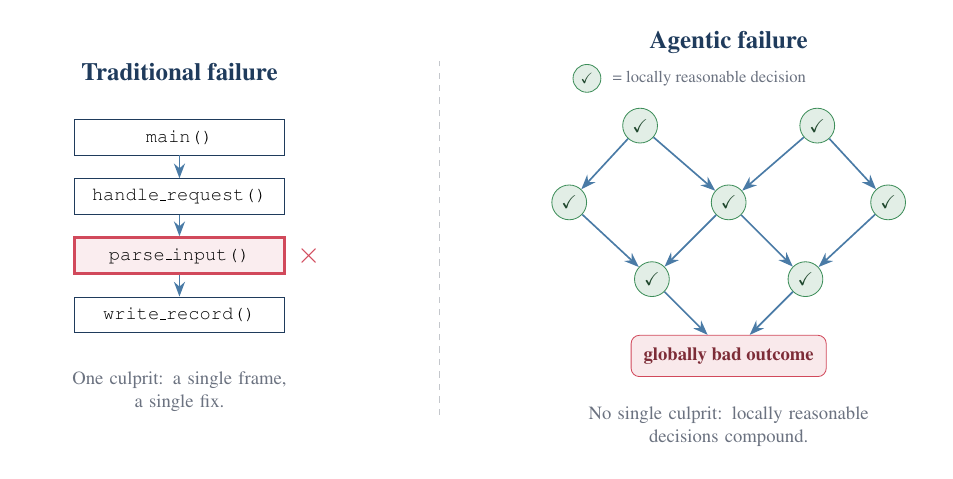}
\caption{Failure attribution. Traditional debugging seeks a single faulty frame in a linear execution trace (left); agentic failures emerge from partial orders of locally reasonable decisions that compound into a globally undesirable outcome (right).}
\label{fig:attribution}
\end{figure}

A third difficulty is failure attribution, which becomes more complex when autonomous systems make decisions through extended sequences of reasoning and action rather than through isolated function calls~\cite{zhang_failure_attribution}.
When a traditional program fails, debugging typically involves identifying a specific faulty function call, failed assertion, or exhausted resource. When an agentic system fails, the cause may be distributed across a sequence of locally reasonable decisions that compound into globally undesirable outcomes~\cite{where_agents_fail,cemri_mas_fail}. Unlike classical planners that offer soundness guarantees, LLM-generated plans may be neither sound nor complete; exploration and refinement may incur significant cost or trigger irreversible state changes before errors become apparent.
Agents that explicitly interleave reasoning and action over extended decision sequences exemplify this challenge~\cite{yao2022react}.
As in distributed systems, reconstructing such failures requires reasoning about partial orderings of events rather than linear execution traces, a challenge long recognized in classical work on distributed causality \cite{lamport_time}. Figure~\ref{fig:attribution} illustrates the contrast.

A related mismatch appears at system interfaces, where long-standing assumptions about control flow and responsibility placement no longer hold.
Microservices architectures assume request–response semantics, while workflow systems assume that control flow is known in advance. Both assumptions reflect an implicit placement of responsibility outside the computational component itself. Agentic systems shift responsibility inward, complicating classical end-to-end reasoning about where correctness, validation, and control should reside \cite{saltzer_end_to_end}.
Agent-initiated requests may also violate implicit assumptions of rate, context, and intent on which services designed for human users rely (Section~\ref{sec:contracts}).

A further class of failures arises at trust boundaries. Computing abstractions have long relied on a separation between code and data: programs are trusted and executed; inputs are processed, not obeyed. Even mechanisms that deliberately blur this line preserve the underlying discipline: an active message selects among handlers installed at the receiver in advance~\cite{voneicken_active}; packets in an active network carry code, but that code runs in a restricted execution environment at each node~\cite{tennenhouse_active}; and mobile code more generally executes only under separately established trust (sandboxes, signatures). Agentic systems collapse the separation into a general problem of \textit{control--data ambiguity}: an agent interprets external content (a web page, an email, the output of another agent) with the same machinery it uses to interpret instructions from its principal, so there is no structural distinction between an observation and a command. Prompt injection, in which adversarially crafted content redirects an agent's behavior~\cite{greshake_injection}, is the currently familiar manifestation, but the underlying question is more durable than any particular attack: \textit{which observations are authorized to modify an agent's goals, plans, beliefs, or action selection?} The resulting failures are not conventional vulnerabilities: no memory-safety violation or privilege escalation occurs, and existing isolation mechanisms are blind to them.

A final class of failures concerns opacity under continuous operation. Interaction models assume explicit commands and episodic execution: a user issues a request, observes the response, and thereby maintains an accurate model of what the system is doing. A continuously operating agent acts between human observations. By the time a problem is noticed, the decision sequence that produced it may be long past, and no human may have been watching when the consequential choices were made. Existing interaction abstractions provide no mechanism for maintaining situational awareness over ongoing autonomous activity~\cite{visibility_agents}. We return to the resulting oversight challenges in Section~\ref{sec:reveals}.

Table~\ref{tab:failure_modes} summarizes these characteristic failure modes of agentic systems and the abstraction gaps they expose. These modes follow from the four properties of Section~\ref{sec:properties}; while systematic empirical study is still developing, each has been documented in recent work on agent benchmarks, deployment experiences, security analyses, and formal treatments~\cite{swebench,kapoor_agents_matter,cemri_mas_fail,agent_contracts,visibility_agents,greshake_injection,anwar_alignment}.

\definecolor{rowgray}{gray}{0.9}
\begin{table}[t]
\centering
\caption{Characteristic failure modes of agentic systems reflect mismatches between persistent, autonomous behavior and existing computational abstractions.}
\vspace{1ex}

\label{tab:failure_modes}
\rowcolors{2}{white}{rowgray}
\begin{tabular}{
>{\raggedright\arraybackslash}p{4.2cm}
>{\raggedright\arraybackslash}p{5.2cm}
>{\raggedright\arraybackslash}p{5.8cm}
}
\hline
\textbf{Observed Failure Mode} &
\textbf{Abstraction Gap} &
\textbf{Resulting Research Challenges} \\
\hline
Autonomous resource consumption and unbounded action initiation &
Execution models assume externally invoked computation with reactive resource control &
Proactive resource budgeting, anticipatory constraint enforcement, isolation for self-directed processes \\[0.5em]

Ambiguous agent lifecycles and costly termination &
Process and service abstractions assume clear start, run, and termination phases &
Lifecycle models for persistent agents, state preservation and transfer, principled shutdown semantics \\[0.5em]

Distributed failure attribution across long decision sequences &
Debugging and verification assume localized faults and linear execution &
Causal tracing across reasoning-action loops, partial-order explanations, long-horizon accountability \\[0.5em]

Misplaced responsibility at system boundaries &
Control and correctness are assumed to reside outside the computational component &
New interface contracts, responsibility assignment, end-to-end reasoning for autonomous initiators \\[0.5em]

Behavior redirected by adversarial content &
Abstractions assume a separation between trusted code and untrusted data &
Trust-boundary tracking, provenance-aware interpretation, limits on untrusted influence over actions \\[0.5em]

Opaque behavior under continuous operation &
Interaction models assume explicit user commands and episodic execution &
Oversight interfaces, escalation policies, scalable human supervision mechanisms \\
\hline
\end{tabular}
\end{table}

\section{Problems the Agency Abstraction Reveals}
\label{sec:reveals}

The failure modes documented above are not independent difficulties to be patched one by one; they form a coherent family of problems that become visible (and tractable) when we take agency as the organizing concept. This section examines these problems across six dimensions (Table~\ref{tab:transitions}), framing each as a contrast between what prior abstractions assumed and what agentic systems require. In each dimension, the required change is an \textit{extension rather than a replacement}: existing work has addressed each concern in specialized settings (budgeted schedulers, contract-rich protocols, temporal verification, supervisory control), but agency makes each concern a general and intrinsic requirement of the computational entity itself.

\subsection{Execution: From Resource Limits to Resource Budgets}

The process abstraction treats resource limits as externally imposed constraints: a process is allocated memory, CPU time, and file handles by the operating system, and operates within these limits or is terminated. The limits do not depend on what the process is trying to accomplish.

Agentic systems invert this relationship. An agent pursuing a goal may need to \textit{negotiate} resource usage based on the value of the goal, the expected cost of achieving it, and the constraints imposed by its principal. A simple limit, such as ``use at most 1~GB of memory,'' does not capture the relevant tradeoffs. An agent might reasonably use more resources for a high-value goal, or voluntarily use fewer when operating under budget pressure.

This shift from limits to budgets exposes new problems:

\paragraph{Budget specification.} How should resource budgets be expressed? Unlike a limit, a budget is a constraint that the agent must reason about. Budgets may be expressed in terms of money (API costs), time (wall-clock deadlines), compute (token limits), or risk (probability of unrecoverable errors). A budget language must allow these heterogeneous constraints to be composed and prioritized.

\paragraph{Anticipatory enforcement.} Traditional resource enforcement is reactive: exceed a limit, and the process is killed. For agentic systems, this may be too late: significant cost may already have been incurred, or irreversible actions taken. Agentic systems require \textit{anticipatory} enforcement: mechanisms that can predict resource usage and intervene before budgets are exhausted. Anticipatory enforcement connects to work on cost prediction~\cite{tsafrir_backfilling,frugalgpt}, early termination~\cite{eet_termination}, and proactive constraint checking.

\paragraph{Graceful degradation.} When a process exhausts its resources, it fails. When an agent approaches budget exhaustion, it should ideally degrade gracefully: simplify its approach, reduce quality, or negotiate for additional resources rather than failing abruptly. Degrading gracefully requires the agent to monitor its own resource consumption and adapt its policy accordingly.

\paragraph{Agent lifecycle.} Budgets extend to existence itself: the process lifecycle (start, run, terminate) assumes that termination is a well-defined event with low cost. For agents, termination may discard valuable accumulated state: learned strategies, contextual knowledge, in-progress plans. Agent lifecycle management requires mechanisms for checkpointing accumulated state, migrating agents between execution contexts, and reasoning about the value of continued operation versus termination~\cite{agentops_survey,kamatar_academy}.

The actor model~\cite{hewitt_actors} provides relevant foundations: actors are isolated, communicate via messages, and can be supervised. But actor semantics concern message handling rather than goal pursuit, and the model does not address resource budgeting for goal-directed behavior.

\subsection{Interfaces: From Protocols to Contracts}
\label{sec:contracts}

Service-oriented abstractions organize interaction around protocols: specified message formats, sequences, and response semantics. A service exposes an API; clients invoke it according to the protocol; the service responds. Responsibility for correct usage lies with the client.

When agents become clients, this division of responsibility breaks down. A service designed for human users may assume requests arrive at human speeds, with human-level contextual awareness, and for human purposes; agent-initiated requests may violate all three assumptions~\cite{visibility_agents}.

The agency abstraction reveals the need for \textit{contracts}, not protocols alone:

\paragraph{Capability specification.} A protocol specifies \textit{how} to invoke a service. A contract must additionally specify \textit{what agents may do}: which actions are permitted, under what preconditions, with what resource implications. Such contracts connect to capability-based security~\cite{dennis_capabilities}, but extend it to encompass goals and intentions. An agent should be able to discover not just what APIs are available, but what behaviors they enable and constrain.

\paragraph{Action effects.} Agents need to reason about the effects of their actions. A contract should specify preconditions (what must be true for the action to succeed), postconditions (what will be true after the action), and invariants (what the action preserves). Such specifications are familiar from design-by-contract in programming languages~\cite{meyer_contract}, but must be extended to handle uncertainty, partial observability, and the possibility that effects depend on environmental state the agent cannot fully observe.

\paragraph{Reversibility.} Not all actions can be undone. Contracts should distinguish reversible actions (which can be rolled back if they prove mistaken) from irreversible ones (which commit the agent to consequences). An agent operating under uncertainty should prefer reversible actions when possible, and require stronger justification for irreversible ones.

\paragraph{Discovery and negotiation.} In open environments, agents cannot know in advance what services are available or what their contracts specify. Agent-service interaction requires mechanisms for discovering available capabilities, negotiating terms of use, and adapting to services that may change over time.

\paragraph{Authority and delegation.} A contract specifies what an agent may do; authority determines on behalf of whom, and in pursuit of which goals, it may do so (Section~\ref{sec:definition}). An agent that delegates a subtask must be able to \textit{attenuate} authority, granting the delegate a subset of its own permissions, budget, and scope, and to revoke it. Classical mechanisms for capability attenuation and least privilege~\cite{dennis_capabilities,saltzer_protection} provide starting points, and distributed systems have long implemented delegation in practice: early delegation architectures~\cite{gasser_delegation}, proxy-credential delegation in Grid security infrastructure~\cite{foster_gsi}, scoped authorization delegation in research automation services~\cite{chard_globus_automation}, and, most recently, authenticated delegation of authority to AI agents~\cite{south_delegation}. The confused-deputy problem~\cite{hardy_deputy}, in which a program exercises its own authority on behalf of a less-privileged requester, reappears whenever an agent acts on content from untrusted sources.

\paragraph{Trust boundaries.} Contracts must also address the provenance of information that crosses interfaces. Because agents interpret data with the same machinery they use to interpret instructions, content flowing in through an interface can influence behavior as if it were a command (Section~\ref{sec:failures}). Contracts should therefore distinguish channels that may legitimately shape the agent's goals and plans from channels that supply data only, and runtimes should track the influence of untrusted inputs on action selection: an information-flow discipline extended from taint tracking to policy generation~\cite{debenedetti_camel}.

The end-to-end argument~\cite{saltzer_end_to_end} observed that functionality placed in the network core cannot fully satisfy application requirements; applications must implement end-to-end checks themselves. For agentic systems, a parallel argument applies: the agent, as the entity with knowledge of its goals, bears ultimate responsibility for achieving them correctly. Contracts support this responsibility by making the terms of interaction explicit, but the agent must still reason about whether invoking a service serves its goals.

\subsection{Correctness: From Functional to Behavioral}
\label{sec:correctness}

Traditional verification asks, in its paradigmatic form: does this program produce correct outputs for its inputs? This is \textit{functional correctness\/}: correctness defined as a relation between inputs and outputs, evaluated at termination. Formal methods has, of course, long addressed reactive and nonterminating systems through temporal logic, refinement, and model checking; the point is not that behavioral specification is new, but that agency makes it the default case rather than the specialist one.

Agentic systems challenge the functional framing in several ways. They may not terminate; they produce behavior over time, not just final outputs; their ``inputs'' (goals, environmental observations) are not fully specified in advance; and their behavior may be correct in some sense even if they fail to achieve their goals, if the failure was reasonable given their information and resources.

The agency abstraction reveals the need for \textit{behavioral correctness}: criteria that evaluate agent behavior over time, relative to goals and constraints.

\paragraph{Safety envelopes.} Some states are unacceptable regardless of the agent's goals. A safety envelope defines a set of states the agent must never enter: states involving irreversible harm, violation of ethical constraints, or catastrophic resource exhaustion. Verification for safety envelopes asks: can we guarantee, regardless of the policy the agent generates, that it will never enter unsafe states? Safety-envelope verification connects to work on safe reinforcement learning~\cite{garcia_safe_rl}, constrained optimization, and formal verification of cyber-physical systems.

We can formalize safety envelopes as follows. Let $S$ be the set of possible states, $A$ the set of actions available through the agent's environment interface $E$, and $H$ the set of interaction histories (finite sequences of observations and actions). For a goal $G$, the policy generator $\Pi$ produces, and may repeatedly revise, a policy $\pi: H \rightarrow A$, conditioned on $G$ and memory $M$. A safety envelope is a set $S_{\text{safe}} \subseteq S$. The agent satisfies the safety envelope if every state reachable under any policy that $\Pi$ may generate lies in $S_{\text{safe}}$.

For open-ended policy generators, verifying this property directly (quantifying over every policy the generator might produce) may be intractable, or achievable only vacuously. That observation points to a systems response rather than a purely logical one: enforce important invariants \textit{outside} the generator, in a smaller trusted mechanism such as a reference monitor, capability system, sandbox, budget governor, or runtime verifier, so that unsafe actions cannot execute regardless of what the generator proposes. This is the classical move of separating policy from mechanism~\cite{levin_hydra} and minimizing the trusted computing base, applied to agents: do not try to prove that the policy generator will never decide to do $X$; construct the substrate so that $X$ cannot occur without invariant $Y$ being satisfied. The verification burden then falls on the small mechanism, not the open-ended generator.

\paragraph{Regret bounds.} An agent that satisfies safety constraints may still perform poorly. Regret measures the gap between the agent's performance and the performance of an optimal policy with full information. A regret bound guarantees that this gap is limited, even though the agent operates under uncertainty. Formally, if $V^*$ is the value achieved by an optimal policy and $V^\pi$ is the value achieved by the agent's policy, regret is $V^* - V^\pi$. Regret bounds are well-studied in online learning and reinforcement learning~\cite{cesa_bianchi_plg,jaksch_ucrl}; the challenge for agentic systems is establishing such bounds for policy generators operating in open environments.

\paragraph{Temporal properties.} Some correctness requirements concern sequences of states over time. Safety properties (``nothing bad ever happens'') and liveness properties (``something good eventually happens'') from temporal logic apply to agent behavior. An agent should never simultaneously hold two conflicting commitments (safety); an agent should eventually respond to every high-priority request (liveness). Temporal logics such as LTL~\cite{pnueli_ltl} and CTL~\cite{clarke_ctl} provide specification languages; the challenge is verification when the agent's policy is not known in advance.

\paragraph{Satisficing criteria.} Following Simon~\cite{simon_bounded}, an agent that cannot achieve optimality may still behave correctly by achieving ``good enough'' outcomes. A satisficing criterion defines a threshold: the agent is correct if it achieves outcomes exceeding this threshold given its resource constraints. Unlike regret bounds (which compare to optimal), satisficing criteria define absolute standards adjusted for context.

\paragraph{Runtime verification.} When static verification is intractable, a practical alternative is to verify at runtime: rather than proving that all possible policies satisfy correctness criteria, runtime monitors observe the agent's behavior and intervene when violations are imminent. This approach requires efficient monitors, clear intervention policies, and mechanisms for the agent to adapt its behavior in response to interventions.

\paragraph{Causal tracing and attribution.} Behavioral correctness criteria are useful only if violations can be diagnosed. When a constraint is violated after an extended decision sequence, attributing the violation requires reconstructing which decisions, observations, and interactions contributed to it. As in distributed systems, the relevant structure is a partial order of causally related events~\cite{lamport_time} rather than a linear trace: an action may depend on an earlier observation, a remembered conclusion, or a message from another agent. Research challenges include recording decision provenance efficiently, constructing partial-order explanations that humans can audit, and maintaining accountability over horizons in which the relevant context has long since left the agent's working memory.

\subsection{Oversight: From Commands to Attention Allocation}

Human-computer interaction has largely been organized around a command model: the human issues commands, the computer executes them. Mixed-initiative interaction and supervisory control have long complicated this picture in specialized settings~\cite{horvitz_mixed}, but even these approaches generally assume episodic interaction with clear turn-taking.

\begin{figure}[t]
\centering
\includegraphics[width=0.62\textwidth]{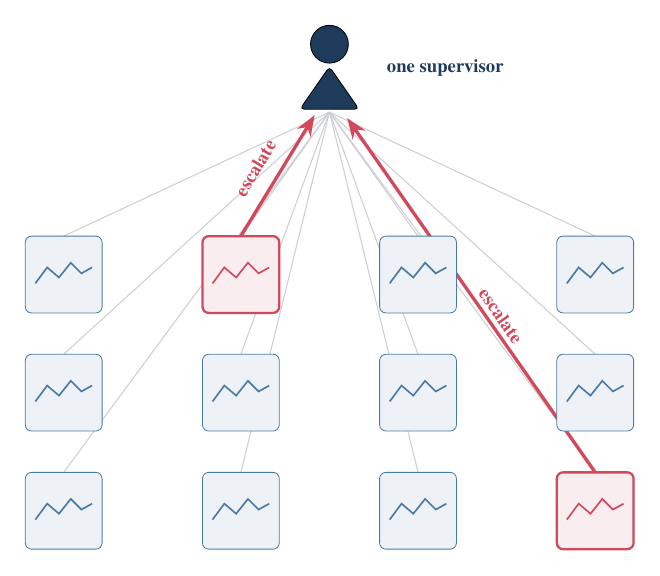}
\caption{Oversight at scale. A single supervisor cannot review every decision of a fleet of continuously operating agents; oversight becomes a problem of allocating scarce human attention, with escalation as a first-class mechanism.}
\label{fig:oversight}
\end{figure}

When agents operate continuously and at scale, the command model fails. A human cannot issue commands to an agent that is always running, pursuing goals that may take hours or days to achieve. The human's role shifts from commanding to \textit{supervising}: monitoring ongoing behavior, intervening when necessary, and allocating limited attention across multiple agents~\cite{anwar_alignment} (Figure~\ref{fig:oversight}). When human attention cannot scale, automated supervisors may assist, raising the further question of how less capable supervisors, human or machine, can oversee more capable systems~\cite{weak_to_strong}.

\paragraph{Attention as scarce resource.} A human supervisor can monitor only a limited amount of agent behavior. With one agent, detailed supervision may be feasible; with hundreds, it is not. The oversight problem becomes one of attention allocation: which agents deserve scrutiny? Which behaviors warrant intervention? Attention allocation connects to work on anomaly detection, alarm management, and supervisory control in other domains.

\paragraph{Escalation policies.} Not all decisions should require human involvement. Effective oversight requires clear escalation policies: criteria that determine when an agent should pause and request human input versus proceeding autonomously. Escalation policies must balance the cost of interruption (delay, attention drain) against the risk of autonomous action (irreversible errors, misaligned behavior). The agent itself may need to reason about whether to escalate, adding another dimension to its policy generation.

\paragraph{Transparency and summarization.} When humans do supervise, they need to understand what agents are doing. Extended decision sequences cannot be presented in full; oversight interfaces must summarize goals, current state, recent actions, and pending decisions at levels of abstraction humans can quickly assess. The requirement is akin to explainable AI~\cite{gunning_xai}, but with the emphasis on temporal summarization of ongoing behavior rather than explanation of individual decisions.

\paragraph{Trust calibration.} Humans form mental models of agent capabilities and adjust supervision accordingly; miscalibrated trust, whether over-trusting a fallible agent or under-trusting a reliable one, leads to oversight failures. Systems should support calibration by providing accurate information about agent capabilities, track record, and confidence.

\subsection{Coordination: From Scheduling to Negotiation}
\label{sec:coordination}

When multiple processes share resources, the operating system \textit{schedules} them: allocating CPU time, memory, and I/O bandwidth according to policies that processes do not control. Market-based allocation and utility-aware schedulers exist, but as specialized designs rather than properties of the process abstraction itself: in the abstraction, processes are scheduled; they do not negotiate.

Multiple agents sharing resources present a different problem. Agents have goals; resource allocation affects goal achievement; agents may have preferences over allocations. Task allocation by negotiation has a classical precedent in the Contract Net Protocol~\cite{smith_contract_net}, in which nodes announce tasks, bid, and award contracts. The shift from scheduling to \textit{negotiation}, and from invocation to \textit{delegation}, exposes new problems:

\paragraph{Delegation with accountability.} An agent may delegate subtasks to  other agents. Unlike procedure calls, delegation between agents involves entities with their own goals that may not perfectly align with the delegator's~\cite{castelfranchi_delegation}. Delegation requires specifying what the delegate should accomplish, what resources it may use, and how accountability flows back to the delegator. When a delegate fails or misbehaves, who is responsible? The compositional question is what must be preserved across delegation: authority, budget, constraints, provenance, accountability, deadlines, revocation rights. Defining delegation semantics that propagate these coherently is, we suggest, the central composition problem for an agent abstraction (Section~\ref{sec:test}).

\paragraph{Conflict resolution.} Agents with different goals may compete for resources or take actions that interfere with each other. Unlike process scheduling, where the scheduler can simply timeslice, agent conflicts may involve goals that cannot all be satisfied. Conflict resolution requires mechanisms for agents to recognize conflicts, negotiate resolutions, and accept outcomes. When negotiation fails, adjudication may be required.

\paragraph{Emergent coordination.} When many agents interact, system-level behaviors may emerge that no individual agent intended~\cite{park_generative}: some beneficial, such as distributed problem-solving, others harmful, such as cascading failures and resource exhaustion spirals. The 2010 Flash Crash, in which interacting automated traders amplified a market collapse, offers a precedent from an earlier generation of autonomous systems~\cite{kirilenko_flash}. Managing emergence connects to work on complex systems, with the added complication that each participant is generating its own policy.

\paragraph{Multi-agent attribution.} When something goes wrong in a multi-agent system, the failure may result from actions of multiple agents, each acting reasonably given its information~\cite{cemri_mas_fail}. Causal analysis must trace how agents' decisions influenced one another, and automatically identifying the responsible agent and the decisive error step remains unreliable even with frontier models~\cite{zhang_failure_attribution}. This problem is akin to fault attribution in distributed systems~\cite{lamport_time}, complicated by the goal-directed nature of agent behavior.

\subsection{Persistence: From State to Memory}

Processes have state; agents require something stronger: \textit{memory}, i.e., persistent state with semantic consequences for future decisions, commitments, and identity. The distinction is not that process state is meaningless; state is often semantically rich and durable. The distinction is that agent memory is state interpreted as part of the agent's continuing identity: its beliefs, experience, commitments, and goals. We first consider the architecture of agent memory, and then problems (provenance, belief revision, forgetting, identity) that are consequences of taking that interpretation seriously. These concerns have counterparts in databases and distributed systems, but agency gives them a different significance: remembered information becomes part of the basis for future autonomous decisions, commitments, and actions.

\paragraph{Memory architecture.} Cognitive science distinguishes working, episodic, semantic, and procedural memory~\cite{squire_memory}; agentic systems exhibit analogous distinctions, with context windows as limited working memory, retrieval augmentation as a form of semantic memory, and fine-tuned behaviors as procedural memory~\cite{sumers_coala}. Which memory architectures support which agent capabilities is an open research direction.

\paragraph{Provenance and trust.} Memories may originate from direct observation, from communication with other agents, or from external sources of varying reliability, and an agent that acts on a memory must weigh its provenance~\cite{buneman_provenance}: a first-hand observation warrants different treatment from a claim reported by an untrusted agent. This requirement parallels the control--data ambiguity of Section~\ref{sec:failures}, but extended in time: adversarial content written into memory can redirect an agent's behavior long after it was encountered~\cite{agentpoison}.

\paragraph{Belief revision.} Agents must update their beliefs when evidence changes. Classical belief revision theory (the AGM framework~\cite{agm}) provides foundations, but applying it to agents whose beliefs may be distributed across explicit memory, retrieved context, accumulated summaries, and model parameters raises new challenges. How should an agent revise beliefs that span such heterogeneous representations, only some of which are explicit propositions?

\paragraph{Forgetting.} Not everything should be remembered forever: storage is finite, old information becomes obsolete, and privacy may require deletion. Agents need mechanisms that remove information gracefully without disrupting ongoing goals: machine unlearning~\cite{bourtoule_unlearning}, in a context where the forgotten information may have influenced current plans and commitments.

\paragraph{Active memory management.} Memory need not be a passive store. An agent may need to decide what it does not know, seek clarification, reconcile conflicting observations, curate what it retains, and judge what information is worth acquiring at what cost. Memory management thereby becomes goal-directed behavior under bounded resources in its own right, extending the progression that organizes this section: from state, to memory, to active knowledge maintenance.

\paragraph{Identity and continuity.} What makes an agent the same agent over time? If its memory is transferred to a new substrate, is it the same agent; if two copies are made, which is the original? These questions, familiar from the philosophy of personal identity~\cite{parfit_reasons}, become practical when agents persist, migrate, and replicate, with direct implications for accountability, commitment, and trust~\cite{chan_ids}.

\section{Implications for Computer Science}
\label{sec:implications}

The problems identified in this paper span subfield boundaries, but share a common root: the mismatch between autonomous, goal-directed computation and abstractions designed for externally controlled execution. This coherence suggests a cross-cutting research agenda for computer science.

\subsection{A Unifying Research Agenda}

Historically, new computational phenomena, from networking to machine learning, have driven integration across subfields. Agentic AI may play a similar integrative role. The failure modes of Table~\ref{tab:failure_modes} do not respect subfield boundaries: a single runaway agent is simultaneously a systems problem, a language problem, a theory problem, and an interface problem, and the dimensions of Table~\ref{tab:transitions} engage, in turn, the systems, programming languages, theory, and human--computer interaction communities.

These problems go beyond applying existing techniques to a new domain. They require rethinking foundational assumptions: that resource bounds are externally imposed, that control flow is specified in advance, that correctness is defined at termination, and that humans command and computers respond.

\subsection{Does Agency Pass Its Own Test?}
\label{sec:test}

Section~\ref{sec:abstractions} defined an organizing abstraction as one that makes problems easy to state, tractable to solve, and natural to compose. We consider each criterion in turn.

\textit{Statement.} Here the case is strongest. Agency exposes resource exhaustion, control--data ambiguity, memory poisoning, and supervision failure as manifestations of a common abstraction gap, autonomous policy generation under constraints, rather than as unrelated engineering problems. The six dimensions of Section~\ref{sec:reveals} are easy to state only once the agent is the unit of analysis.

\textit{Tractability.} Once the agent is an explicit system entity, a runtime can associate budgets, capabilities, authority, provenance, identity, commitments, and monitors with that entity, and enforce invariants around the policy generator through small trusted mechanisms (Section~\ref{sec:correctness}). This arrangement parallels how the process abstraction lets an operating system associate address spaces and resource limits with running programs. Here the evidence is a research agenda rather than an accomplished fact: the constituent mechanisms exist in classical systems work, but their integration around a policy generator remains to be built.

\textit{Composition.} A mature abstraction has composition rules: processes compose through interprocess communication, objects through interfaces, services through protocols, and transactions through transactional semantics. Agents compose through \textit{delegation}, and the composition rule is the open question identified in Section~\ref{sec:coordination}: what must be preserved when one agent delegates a goal to another---authority, budget, constraints, provenance, accountability, revocation? An agent abstraction will be mature when delegation has defined semantics for propagating these properties.

\begin{figure}[t]
\centering
\includegraphics[width=0.72\textwidth]{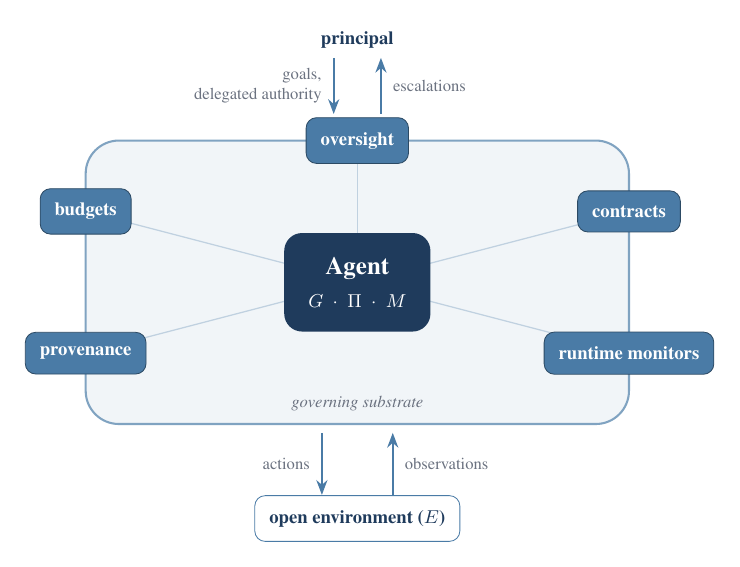}
\caption{The agent abstraction in synthesis. The agent of Figure~\ref{fig:agent_architecture}, now exercising authority delegated by its principal, embedded in a governing substrate whose mechanisms (budgets, contracts, provenance, runtime monitors, oversight) make generated behavior governable.}
\label{fig:synthesis}
\end{figure}

Figure~\ref{fig:synthesis} summarizes the resulting architecture: a computational entity defined by goals, a policy generator, and memory, exercising delegated authority within a substrate of mechanisms (budgets, contracts, provenance, monitors, oversight) that make generated behavior governable. Agency is thus a candidate abstraction: strongest today on the first criterion, and a concrete research program on the other two.

\subsection{Research Priorities}

The problems constituting this research program vary in urgency and tractability. We suggest a rough prioritization:

\paragraph{Near-term.} Some problems affect currently deployed systems and are tractable extensions of existing techniques: execution environments with budget negotiation, runtime monitoring for behavioral properties, trusted enforcement mechanisms around policy generators, and oversight interfaces for attention allocation.

\paragraph{Medium-term.} Other problems require more fundamental development but build on established foundations: programming models for constrained autonomy, formal foundations for satisficing correctness, and delegation semantics for authority and budgets.

\paragraph{Long-term.} Some problems are not yet well enough understood to address directly: theoretical frameworks for emergent behavior in multi-agent systems, principled approaches to agent identity and continuity, and belief revision in neural systems. Progress here may require new conceptual frameworks rather than application of existing ones.

\subsection{Technical-Governance Interactions}

Technical and governance challenges interact in important ways. By \textit{governance} we mean the external processes by which organizations and societies direct and hold accountable the behavior of deployed systems: internal policies, audits, certification, incident response, and regulation~\cite{shavit_governing,kolt_governing}. Well-designed abstractions can make governance more tractable by providing hooks for auditing, constraint enforcement, and oversight. A runtime that tracks resource usage, logs decisions, and enforces budgets provides natural points for regulatory inspection. A contract language that makes agent capabilities explicit enables meaningful certification.

Conversely, poorly designed abstractions can make governance harder by obscuring behavior or creating fait accompli through deployed systems that resist modification.

Technical researchers therefore have responsibilities extending beyond narrow efficiency metrics. The abstractions we develop will shape what governance is possible. Designing abstractions with governance in mind (building in logging, auditability, constraint enforcement, and human override) is not an optional addition but an integral part of the research agenda.

\subsection{Path Dependence in Infrastructure}

Abstractions have architectural consequences: whichever primitives become conventional will determine where resource control, identity, authority, provenance, audit, and responsibility reside. If today's agent frameworks establish conventions in which authority is represented only implicitly in prompts, memory carries no explicit provenance, and resource constraints are enforced only after actions are generated, those choices will become embedded in APIs, protocols, and ecosystems, much as the socket, the file descriptor, and the container embedded earlier design choices. Retrofitting first-class authority, budgets, or provenance into an installed base is far harder than designing them in. An explicit agency abstraction is an opportunity to make these properties first class while the conventions are still forming.

\section{Related Work}
\label{sec:related}

Our thesis invites objections of the form ``this is not new.'' We organize related work around such objections, asking in each case what the earlier tradition established and what changes in the setting considered here.

\paragraph{``Agents are not new.''} True: classical agent theory~\cite{russell_norvig}, multi-agent systems research~\cite{wooldridge_agents}, and belief-desire-intention (BDI) architectures~\cite{rao_georgeff} established agents that perceive, act, deliberate, and coordinate decades ago, and they supply much of the vocabulary used here. What these traditions did not do, because the enabling capability did not exist, was treat the agent as a \textit{systems} abstraction with lifecycle, budget, authority, and oversight semantics.

\paragraph{``Autonomous systems are not new.''} Autonomic computing~\cite{kephart_autonomic}, autonomous spacecraft~\cite{muscettola_remote_agent}, robotics, and adaptive control have long operated without step-by-step human direction, and our discussions of lifecycle management and graceful degradation draw on them. As argued in Section~\ref{sec:policygen}, however, these systems execute or adapt policies within structures fixed at design time; the regime addressed here is the runtime generation of behavior over incompletely modeled environments.

\paragraph{``Dynamic policy generation is not new.''} Automated planning~\cite{ghallab_planning}, program synthesis~\cite{gulwani_synthesis}, and reinforcement learning all construct behavior rather than selecting it from a fixed repertoire. The distinction, developed in Section~\ref{sec:policygen}, lies at the action--model boundary: these methods presuppose explicit state and action models constructed before the system runs, whereas contemporary agents act where models are incomplete, discovered dynamically, or represented implicitly in a foundation model.

\paragraph{``Persistent distributed entities are not new.''} The actor model~\cite{hewitt_actors} is perhaps the closest prior organizing abstraction: isolated, persistent entities communicating by message. But actor semantics specify how messages are handled, not how goals are formulated and pursued, and the model offers no account of budgets or authority. Services~\cite{erl_soa} and Grid virtual organizations~\cite{foster_anatomy} added distribution, interoperability, and cross-domain trust; early calls for Grid--agent convergence~\cite{foster_brain_brawn} articulated an agenda that this paper revisits (Section~\ref{sec:abstractions}) in light of foundation models.

\paragraph{``The safety and oversight problems are not new.''} Formal verification of reactive systems~\cite{pnueli_ltl,clarke_ctl}, safe reinforcement learning~\cite{garcia_safe_rl}, AI safety and alignment~\cite{amodei_safety}, and research on supervisory control and mixed-initiative interaction~\cite{horvitz_mixed} all address closely related problems. We draw on each. What changes is the object and the scale: verification and supervision must now target a policy generator rather than a policy, and must extend across populations of continuously operating entities. Better abstractions will not solve alignment, but whatever alignment approaches prove effective will require them as infrastructure.

\paragraph{``Agent frameworks already supply this.''} A rapidly growing ecosystem of frameworks supports the construction of LLM-based agents, including tool-use mechanisms~\cite{schick_toolformer}, reasoning--action loops~\cite{yao2022react}, and multi-agent conversation frameworks~\cite{wu_autogen}, alongside widely used engineering platforms. These efforts supply valuable mechanisms, but they are engineering frameworks rather than organizing abstractions: each makes local design decisions about memory, control, and tool invocation without providing the principled foundations (lifecycle semantics, budget enforcement, behavioral correctness criteria, oversight interfaces) whose absence this paper documents. The closest conceptual neighbor is the proposal of cognitive architectures for language agents~\cite{sumers_coala}, which organizes the agent's \textit{internal} structure (memory, action space, decision-making) from a cognitive-science perspective; the present paper is complementary, treating the agent as a \textit{systems} abstraction with external semantics: lifecycle, budgets, authority, and oversight. One goal of articulating agency as an abstraction is to give frameworks and architectures alike a common conceptual target.

\section{Conclusion}
\label{sec:conclusion}

We have argued that agency merits recognition as an organizing abstraction for computer science: a conceptual framework that structures how we build, reason about, and verify a class of computational systems.

The core claim is a reframing of what computation is organized around: \textit{a process executes a program; an agent pursues a goal}. The process abstraction organizes computation around executing code specified in advance; the agency abstraction organizes computation around pursuing goals under bounded resources, by generating and revising courses of action under constraints of information, resources, authority, and time. This shift reveals a family of related problems, spanning execution environments, interface design, correctness criteria, human oversight, coordination, and persistence, that existing abstractions address only in specialized settings and that agency makes general and intrinsic.

The distinctive feature of contemporary agentic systems, built on foundation models, is \textit{open-ended policy generation}: unlike prior autonomous systems that execute predefined policies or synthesize behavior within models specified in advance, LLM-based agents generate courses of action at runtime that did not exist before the goal was given. The same capability that makes them powerful makes them difficult to govern.

We have identified specific problems in each of the six dimensions of Table~\ref{tab:transitions}, sketched the new mechanisms that the agency abstraction requires, and argued that many important invariants are best enforced by small trusted mechanisms surrounding the policy generator rather than by verifying the generator itself.

Whether agency ultimately joins procedures, objects, processes, and services in the canon of durable organizing abstractions remains to be seen. The underlying gap, however, seems unambiguous: computer science has powerful abstractions for controlling the execution of computations specified in advance, and lacks comparably mature abstractions for controlling computations that decide at runtime what behavior to generate. Supplying them is the research program this paper proposes. Deployed agentic systems are already encountering the challenges we have identified; the program is not only necessary but urgent.

We do not claim that better abstractions alone will resolve the profound questions of governance, accountability, and social impact that agentic AI raises~\cite{singapore_consensus}. We do claim that well-designed abstractions are necessary infrastructure: they provide the hooks for auditing, the mechanisms for constraint enforcement, and the interfaces for oversight that governance requires. By engaging these technical problems directly, while maintaining appropriate humility about their scope, computer science can contribute to a more principled foundation for autonomous computational systems.

\section*{Acknowledgments}

I thank the anonymous \textit{Communications of the ACM} reviewers, whose comments substantially improved the article on which this expanded version is based, and the many colleagues with whom I have discussed these topics.
This work was supported in part by the U.S.\ Department of Energy, Office of Science, under Contract No.\ DE-AC02-06CH11357, and the U.S.\ National Science Foundation, under grant NSF~2004894.

\printbibliography

\end{document}